\documentclass[sigconf, final]{acmart}
\AtBeginDocument{%
  \providecommand\BibTeX{{%
    \normalfont B\kern-0.5em{\scshape i\kern-0.25em b}\kern-0.8em\TeX}}}

\setcopyright{acmcopyright}
\copyrightyear{2018}
\acmYear{2018}
\acmDOI{10.1145/XXXXXXX}

\acmConference[ACM Multimedia '25]{Dublin '25: ACM Multimedia}{27-31 October 2025}{Dublin, Ireland}
\acmPrice{15.00}
\acmISBN{978-1-4503-XXXX-X/18/06}

\acmSubmissionID{73}

\usepackage{times}
\usepackage{epsfig}
\usepackage{graphicx}
\usepackage{textcomp}
\usepackage{tikzducks}

\usepackage{algorithm, algpseudocode}
\usepackage{multirow}
\usepackage{caption}
\usepackage{comment}
\definecolor{citecolor}{RGB}{119,185,0} 

\usepackage{pifont}
\usepackage{color}
\usepackage{cleveref}

\newlength\savewidth

\usepackage{acro}

\DeclareAcronym{cvgl}{
    short = CVGL,
    long  = Cross-View Geo-Localisation,
    tag   = nomencl
}
\DeclareAcronym{gnss}{
    short = GNSS,
    long  = Global Navigation Satellite Systems,
    tag   = nomencl
}
\DeclareAcronym{fov}{
    short = FOV,
    long  = Field-of-View,
    tag   = nomencl
}
\DeclareAcronym{llm}{
    short = LLM,
    long  = Large Language Model,
    tag   = nomencl
}
\DeclareAcronym{llms}{
    short = LLMs,
    long  = Large Language Models,
    tag   = nomencl
}

\DeclareAcronym{vlm}{
    short = VLM,
    long  = Vision-Language Model,
    tag   = nomencl
}
\DeclareAcronym{vlms}{
    short = VLMs,
    long  = Vision-Language Models,
    tag   = nomencl
}

\newcommand{\topone}{30.21}

\newcommand{\topten}{63.13}

\newcommand{\papername}{VICI}

\usepackage{fancyvrb}
\usepackage{fvextra} 

\begin{document}
\title{VICI: VLM-Instructed Cross-view Image-localisation}





\author{Xiaohan Zhang$^{1}$* \quad Tavis Shore$^{2}$* \quad Chen Chen$^{3}$ \quad Oscar Mendez$^{4}$ \\ Simon Hadfield$^{2}$ \quad Safwan Wshah$^{1}$ } 
\affiliation{
  \institution{University of Vermont$^{1}$ \quad University of Surrey$^{2}$ \quad University of Central Florida$^{3}$ \quad Locus Robotics$^{4}$}
  }
\thanks{*Denotes equal author contribution.}

\email{xiaohan.zhang@uvm.edu,   t.shore@surrey.ac.uk
}

\begin{abstract}
In this paper, we present a high-performing solution to the UAVM 2025 Challenge~\cite{wang2025UVA}, which focuses on matching narrow \ac{fov} street-level images to corresponding satellite imagery using the University-1652 dataset. 
As panoramic Cross-View Geo-Localisation nears peak performance, it becomes increasingly important to explore more practical problem formulations. 
Real-world scenarios rarely offer panoramic street-level queries; instead, queries typically consist of limited-FOV images captured with unknown camera parameters. 
Our work prioritises discovering the highest achievable performance under these constraints, pushing the limits of existing architectures. 
Our method begins by retrieving candidate satellite image embeddings for a given query, followed by a re-ranking stage that selectively enhances retrieval accuracy within the top candidates.
This two-stage approach enables more precise matching, even under the significant viewpoint and scale variations inherent in the task.
Through experimentation, we demonstrate that our approach achieves competitive results -
specifically attaining R@1 and R@10 retrieval rates of \topone\% and \topten\% respectively. 
This underscores the potential of optimised retrieval and re-ranking strategies in advancing practical geo-localisation performance. \newline Code is available at \href{https://github.com/tavisshore/VICI}{github.com/tavisshore/VICI}.
\end{abstract}


\maketitle

\section{INTRODUCTION}
Localisation is a fundamental requirement in mobile robotics, as agents must ascertain their position before executing assigned tasks.
These generally rely on \ac{gnss} for localisation; however, this approach becomes unreliable in urban canyons or conflict zones, where signal obstruction, multipath effects, or deliberate jamming degrade performance.
\ac{cvgl} offers a robust alternative to address this issue by inferring the location of a street-level image by matching it to a corresponding geo-tagged satellite image in which it appears.
In most existing works~\cite{geodtr,geodtr+,sample4geo,Shi2020WhereAI,SAIG}, limited \ac{fov} ground query images are not fully explored due to the extreme lack of contextual surrounding information. 
Instead, existing methods primarily focused on panoramic imagery, leveraging its wide \ac{fov} to extract descriptive features and optimise matching accuracy. 
However, the majority of mobile robots, from autonomous vehicles to warehouse platforms, are equipped with limited-\ac{fov} cameras, thereby impeding the practical deployment of such systems~\cite{zhang2023cross,wilson2024image}. 
But, directly applying such a feature-matching paradigm to limited-\ac{fov} images often introduces noise, making features less distinguishable, leading to a reduction in \ac{cvgl} performance~\cite{mi2024congeo,shi2020looking}.

\begin{figure}[t!]
    \centering
    \includegraphics[page=1, width=\columnwidth, trim={0 10.4cm 0 0}, clip]{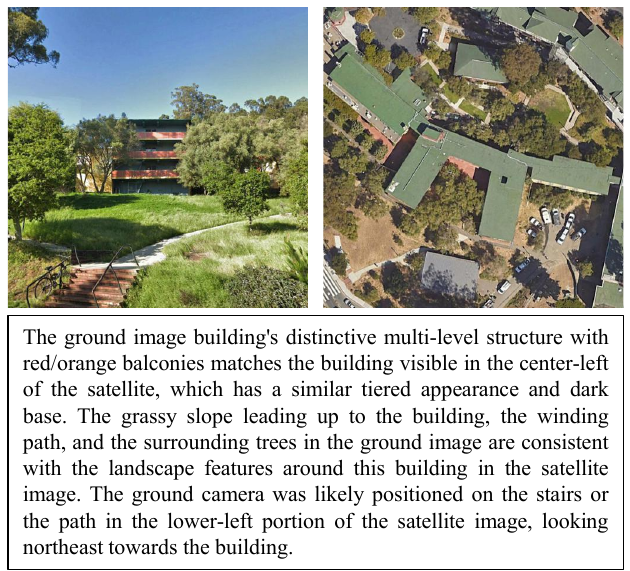}
    \caption{localisation example from \papername{}. \textit{Top left:} query image. \textit{Top right:} top retrieved satellite image. \textit{Bottom:} 
    justification for this satellite image being re-ranked to Top-1.
    }
    \label{fig:teaser_image}
    \vspace{-20pt}
\end{figure}

The recent emergence of \ac{llms} and \ac{vlms}, including ChatGPT~\cite{chatgpt}, Gemini~\cite{team2023gemini}, and LLaMA~\cite{touvron2023llama}, has showcased the strength of these foundation models in image understanding~\cite{liu2023visual}, visual question answering~\cite{VQA}, and text-to-image synthesis~\cite{tian2024visual}.
One idea to alleviate the above-mentioned problem is to leverage the reasoning capability of such models to match the query image against the reference satellite database, providing justification for the matching and offsetting the loss of information. 
However, naively applying \ac{vlms} on the whole reference database is costly and inefficient. To tackle this issue, we propose \papername{}, \textbf{V}LM-\textbf{I}nstructed \textbf{C}ross-view \textbf{I}mage-localisation, a novel two-stage VLM-powered \ac{cvgl} model. 
The first stage extracts visual features from both ground and satellite views - predicting a coarse ranking for the ground query. To alleviate the over-fitting issue, we incorporate drone images to augment the satellite data. 
In the next stage, the Top-10 retrieved candidate satellite images are re-ranked by a \ac{vlm}, which also takes the query image and a curated prompt as input.
In this manner, \papername{} not only improves the localisation accuracy but also maintains the computational overhead at a reasonable scale. 
Furthermore, \papername{} not only re-ranks the predictions but also provides the justifications for the re-ranking decisions. One localisation example with justifications is shown in~\Cref{fig:teaser_image}.

We achieve very competitive performance in the challenge~\cite{wang2025UVA}, attaining R@1 and R@10 retrieval rates of \topone\% and \topten\% respectively. \\

\noindent
In summary, our research contributions are:
\begin{itemize}
    \item Introduction of \papername{}, a novel two-stage \ac{cvgl} framework that integrates \ac{vlm}s to go beyond traditional feature similarity methods. Our approach not only substantially improves localisation accuracy but also introduces interpretable reasoning through language-based justifications.
    \item A novel data augmentation technique that incorporates high-angle drone imagery within the satellite image branch to enhance model robustness and generalisation.
    \item Extensive experiments demonstrate the competitive performance of our \papername{} on the UAVM 2025 challenge~\cite{wang2025UVA}. 
    We also provide quantitative and qualitative evidence for the superiority of the novel two-stage design, illustrating a potential new research direction for the field of \ac{cvgl}.
\end{itemize}

\section{RELATED WORKS}

\noindent\textbf{Cross-View Geo-Localisation: } The deep learning era of \ac{cvgl} began with Workman and Jacobs \cite{7301385}, who demonstrated the efficacy of Convolutional Neural Networks (CNNs) for correlated feature extraction across different viewpoints.
CVGL datasets primarily consisted of panoramic street-level and satellite image pairs, including CVUSA \cite{cvusa}, CVACT \cite{liu2019lending}, and VIGOR \cite{zhu2021vigor}. Recognising the need to better model real-world scenarios, Shi et al. \cite{Shi2020WhereAI} introduced the limited Field-of-View (FOV) crops into the \ac{cvgl} research. Shore et al. \cite{10943342} proposed representing data as a graph, leveraging connectivity information to enhance performance, and subsequently \cite{peng} increasing discriminability by adding reference street-level images to this representation. More recently, to improve overall generalisation and address limited dataset diversity, Huang et al. released CV-Cities \cite{huangCVCities2024}, encompassing a wider range of global city scenes.

Backbone feature extractors play a vital role in \ac{cvgl}. Recently, transformers \cite{vaswani2017attention} were introduced to \ac{cvgl} by two seminal works\cite{Yang2021CrossviewGW, Zhu2022TransGeoTI}. Yang et al. \cite{Yang2021CrossviewGW} combined a ResNet backbone with a vanilla ViT encoder. Zhu et al. \cite{Zhu2022TransGeoTI} proposed a transformer that uses an attention-guided non-uniform cropping to remove uninformative areas.
Zhu et al. \cite{SAIG} introduced an attention-based backbone, representing long-range interactions among patches and cross-view relationships with multi-head self-attention layers.
Sample4Geo \cite{sample4geo} proposed two sampling strategies, sampling geographical sampling and hard sample mining to improve \ac{cvgl} accuracy.
In GeoDTR \cite{geodtr, geodtr+}, Zhang et al. decouple geometric information from raw features, learning spatial correlations within visual data to improve performance. 

\begin{figure}[t!]
    \centering
    \includegraphics[width=0.98\columnwidth, trim={0 3.4cm 0 0}, clip]{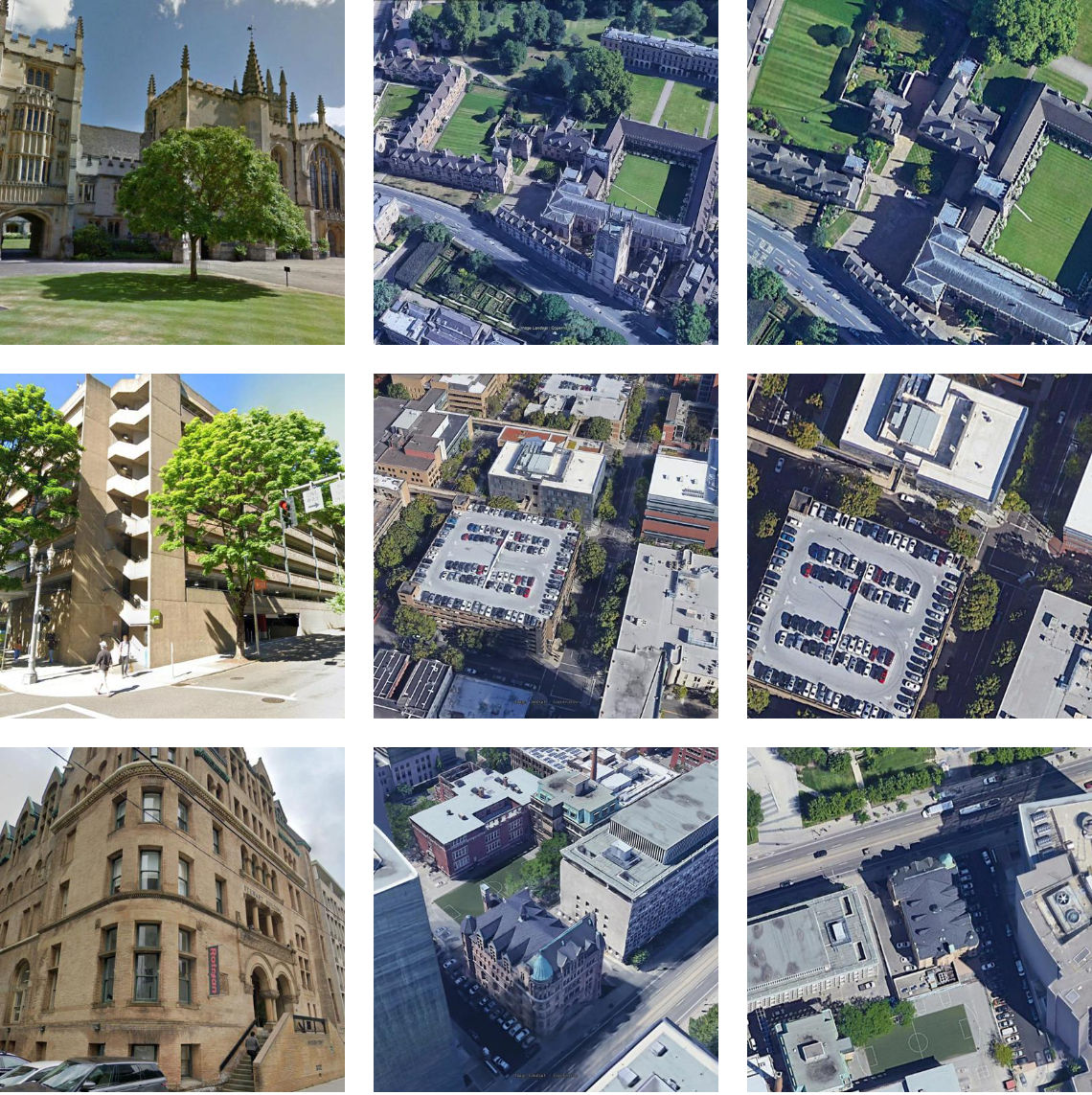}
    \caption{Street-level, drone, and satellite images from various locations, illustrating how the drone imagery provides feature continuity between the viewpoints.}
    \label{fig:aug}
    \vspace{-18pt}
\end{figure}

\noindent\textbf{Vision-Language Models for Geo-localisation:} \ac{vlms} are increasingly being used in image localisation for their logical reasoning capabilities.
Initially, they were fine-tuned to operate with street-level images, combining viewed features to logically determine location.
GeoReasoner \cite{li2024georeasoner} is a two-stage fine-tuned large \ac{vlm} that mimics human reasoning from geographic clues to accurately predict locations from street-level images.
Ye et al. \cite{ye2024cross} introduce a text-guided \ac{cvgl} method that retrieves satellite images using natural language descriptions of street-level scenes, enabling localisation without requiring a query image.
Dagda et al. propose GeoVLM \cite{geovlm}, using a Vision-Language Model to perform zero-shot \ac{cvgl} by re-ranking candidate satellite-ground image pairs using language-based scene descriptions. 
Whereas \cite{geovlm} employs VLMs to categorise specific features within images, our approach leverages VLMs to directly compare image pairs, enabling them to re-rank retrievals and provide justification for their decisions.

\begin{figure*}[tbp!]
    \centering
    \includegraphics[width=0.9\textwidth]{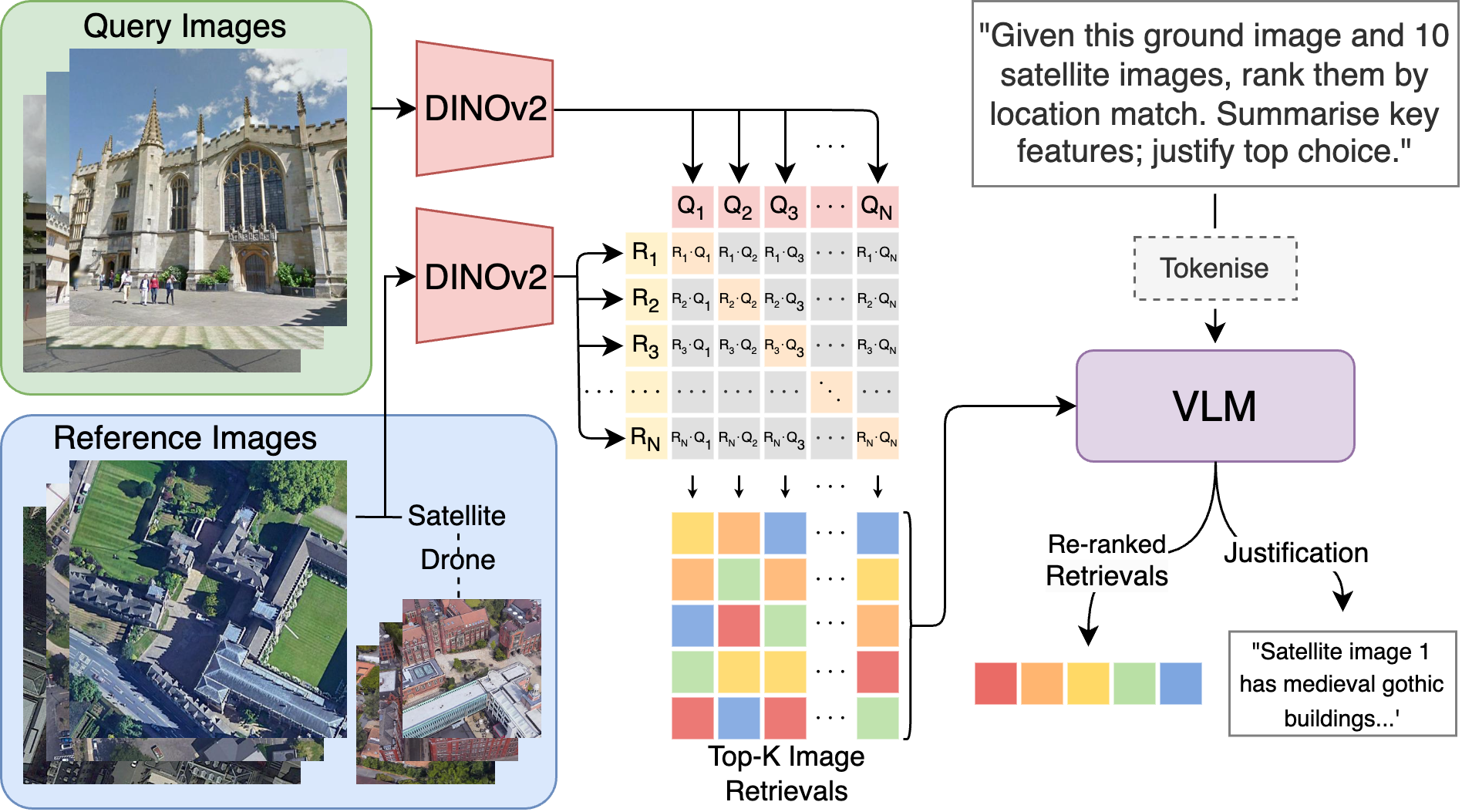}
    \caption{Overview of the architecture: Features extracted by separate DINOv2 branches, references retrieved by descending similarity, re-ranked and justified by passing through a VLM with a text prompt.}
    \label{fig:method_diagram}
    \vspace{-13pt}
\end{figure*}

\section{METHODOLOGY}
The proposed method, \papername, operates in two stages: 1) Coarse Retrieval, extracting feature embeddings from input images - ranking reference embeddings by similarity to the query, and 2) VLM Re-ranking and Reasoning,  re-ranking the candidates from step 1) using a VLM prompted to focus on salient static features shared across viewpoints. 
An overview of the proposed technique is shown in Figure \ref{fig:method_diagram}.

\vspace{-0.5em}
\subsection{Stage I: Coarse Retrieval}
In Stage I, we employ a Siamese network $f$, without weight sharing, parametrised by $\theta$ to minimise the domain gap between street-level and satellite image features, generating embeddings $\eta_{t}$ where $t \in \{street, sat\}$. 
All input images $I$ are in RGB space: $I \in \mathbb{R}^{3{\times}W{\times}H}$, and are resized to $W \times H, \text{\textit{where} } H = W, W \in \{224, 384, 448, 518\}$ for different backbone extractor configurations. Mathematically, this process can be defined as follows, \vspace{-0.5em}

\begin{equation}
    \eta_t = f_\theta\left(I_t \right), t \in \{street, sat\}
    \label{eq:feat_street}
\end{equation}

\noindent During inference, reference embeddings are precomputed and stored offline, querying this database for retrievals during online operation.

\noindent\textbf{Drone Perspective Augmentation: } The challenge dataset, University-1652~\cite{zheng2020university}, curates street, drone, and satellite views. Although this challenge focuses on street-to-satellite \ac{cvgl}, drone imagery provides intermediate, low-altitude oblique views that are significantly more similar to street-level imagery than traditional nadir satellite views. 
More specifically, unlike satellites that observe scenes from near-vertical angles at high altitudes, drones capture structures and terrain from oblique angles and much lower altitudes—typically tens to hundreds of metres—making them closer in both scale and viewpoint to ground-based images (as shown in Figure \ref{fig:aug}).
Thus, drone-view images can assist in bridging the domain gap between ground and satellite viewpoints.

Inspired by this, during training, we randomly feed drone-view images into the satellite branch in place of the satellite images according to a probability $P$, alongside their corresponding street-level images. 
Experimentation demonstrates how the geometric and visual similarities between drone and ground views help narrow the domain gap between aerial and ground perspectives, enabling more effective feature matching and correspondence estimation.
The improved alignment in perspective mitigates issues like occlusion, foreshortening, and extreme viewpoint disparity. 
As a result, drone-satellite fusion enhances spatial reasoning and improves \ac{cvgl} accuracy, particularly in dense urban or structurally complex scenes.

\subsection{Stage II: VLM Re-ranking and Reasoning}
\ac{vlms} recently demonstrated strong performance in recognising objects, interpreting scenes, and aligning visual content with natural language~\cite{team2023gemini,liu2023visual}.
In the second stage of \papername, we leverage this capability by feeding the top-10 retrievals from the reference database into a VLM along with a curated prompt and the query street-level image. 
This prompts the VLM to logically reason about the candidates, produce a more accurate ranking, and justify the decisions.
Below is a simplified version of the text prompt used to re-rank the retrieved satellite images:

\begin{Verbatim}[breaklines=true, breaksymbolleft={}, breaksymbolright={}, fontsize=\small]
Given one ground image and 10 satellite images, identify which satellite image matches the ground location. Summarise the ground image and each satellite image, focusing on key features (streets, buildings, etc.). Then, compare the ground image with each satellite image as well as the summarisation. Rank these 10 satellite images by likelihood [1–10]. Justify the top choice with matching features and estimated camera position.
\end{Verbatim}

\noindent After receiving the response, we extract the re-ranked results from the VLM output, along with the justification for the top choice.


\subsection{Implementation Details}
\noindent\textbf{Stage I:} The stage I of \papername{} is implemented in PyTorch~\cite{pytorch} and trained with InfoNCE loss~\cite{sample4geo} for 100 epochs and batch size of 32 using an AdamW~\cite{adamw} optimiser with an initial learning rate of 1e-5 and an exponential scheduler with gamma of $0.9$. 
We employed a wide variety of backbone feature extractors such as ConvNext~\cite{convnext}, ViT~\cite{ViT}, and DINOv2~\cite{dino}. Training and testing of Stage I are conducted on 4 AMD MI210 accelerators. 

\noindent\textbf{Stage II:} The second stage of \papername{} for re-ranking and reasoning is performed with Google's Gemini 2.5 \cite{google2025gemini25flash}. 
We chose two model variants, Gemini 2.5 Flash and Gemini 2.5 Flash Lite, to balance efficiency and cost. 
We set the temperature to 0 to have a fixed output and better reproducibility. 
We leverage the structured output functionally~\footnote{https://ai.google.dev/gemini-api/docs/structured-output} to make the output follow a JSON structure. 
For more details, please refer to the corresponding code.

\section{EVALUATION}
\label{sec::eval}

\noindent\textbf{Dataset:} This challenge~\cite{wang2025UVA} utilises the University-1652 dataset \cite{zheng2020university} for benchmarking purposes. 
This dataset contains image sets of 1,652 unique university buildings: 701 for training and the rest for testing.
Each image set contains a single satellite image featuring the building, $54$ drone images captured with an ascending circling trajectory, and a few street-level, limited-FOV images cropped from street view panoramas. All images share the same resolution of $512 \times 512$. 
To have a fair comparison and illustrate the power of the proposed \papername{} under the case of limited training data, we did \textbf{NOT} include extra training data, although it is allowed in this challenge.

\noindent\textbf{Backbone Comparison:} The first experiment aims to identify the most effective feature extraction architecture for this challenging limited-FOV \ac{cvgl} task.
We conducted a comprehensive evaluation with uniform training conditions. 
The results are summarised in~\Cref{tab:back_eval}. 
Although ConvNeXt has demonstrated promising results in previous work~\cite{sample4geo,geodtr+}, its performance on this challenging limited-FOV dataset is the weakest among all evaluated backbone architectures.
This may be due to the limited capacity of ConvNeXt to capture sufficient contextual information from narrow FOV images.
We then experiment with two large-scale backbones, ViT and DINOv2, observing that even with similar model structures, ViT~\cite{ViT} is consistently worse than DINOv2~\cite{dino}. 
Interestingly, with almost the same structure, DINOv2 consistently performs better than DINOv2 on Base scale (``B'') and Large scale (``L''). 
This performance increase might result from 1) the pre-trained knowledge of DINOv2 on the LVD-142M dataset~\cite{dino}, and 2) the native input image size contains more fine-grained details (as illustrated by the different FLOPs). 
From this study, we selected DINOv2-L as the stage I backbone for \papername.

\begin{table}[h!]
\centering
\resizebox{\columnwidth}{!}{%
\begin{tabular}{cccc|ccc}
Backbone & Params (M) & FLOPs (G) & Dims & R@1 & R@5 & R@10 \\ \hline
ConvNeXt-T & 28 & 4.5 & 768 & 1.36 & 4.34 & 7.95  \\
ConvNeXt-B & 89 & 15.4 & 1024 & 3.14 & 8.14 & 13.22 \\
ViT-B & 86 & 17.6 & 768 & 3.30 & 8.92 & 13.96 \\
ViT-L & 307 & 60.6 & 1024 & 9.62 & 23.42 & 32.73 \\
DINOv2-B & 86 & 152 & 768 & 17.37 & 36.14 & 46.96  \\
DINOv2-L & 304 & 507 & 1024 & 27.49 & 51.96 & 63.13
\end{tabular}%
}
\caption{
Backbone capabilities evaluation.
}
\label{tab:back_eval}
\vspace{-15pt}
\end{table}

\noindent\textbf{Drone Perspective Augmentation:} The second experiment evaluates the effectiveness of the drone perspective augmentation, as summarised in~\Cref{tab:drone_aug}.
For each training sample, we design a probability $P$ to replace the satellite image with a randomly sampled drone-view image from the same location. 
In this experiment, we set $P$ to $0$, $0.1$, $0.3$, and $0.5$, respectively. 
As we can see, by setting $P$ to $0.1$ and $0.3$, the model performance significantly improves - with $P=0.3$ achieving the best results. 
However, with $P=0.5$, performance drops significantly and is similar to the non-augmented case. 
Thus, we choose $P=0.3$ for the probability of drone perspective augmentation during the training.

\begin{table}[h!]
\centering
\setlength{}{}
\setlength{\tabcolsep}{12pt}
\begin{tabular}{c|ccc}
$P$ & R@1 & R@5 & R@10 \\ \hline
0 & 24.47 & 48.16 & 60.99  \\
0.1 & 26.98 & 51.34 & 61.92 \\
0.3 & 27.49 & 51.96 & 63.13 \\
0.5 & 24.89 & 52.03 & 62.66
\end{tabular}%
\caption{Drone augmentation with varying probability $P$.}
\label{tab:drone_aug}
\vspace{-20pt}
\end{table}

\noindent\textbf{VLM Re-ranking:} 
The top 10 retrieved results for each query ground image are fed into stage II for VLM Re-ranking. 
To balance computational cost and efficiency, we choose two different variants of Google's Gemini 2.5 model - Gemini 2.5 Flash and Gemini 2.5 Flash Lite. 
We also fix the thinking budget at $1024$ for both models to achieve the best efficiency. Results are summarised in~\Cref{tab:vlm_eval}. By comparing the results with and without re-ranking utilising Gemini 2.5 Flash, R@1 increases by 2.72\% and R@5 increases by 1.08\%, supporting the idea of leveraging \ac{vlms} to re-rank the coarse retrieving results. 
To further investigate the \ac{vlm} re-ranking performance on small-scale models, we conducted the same experiment on the Gemini 2.5 Flash Lite. 
However, performance is worse than without re-ranking, illustrating that large-scale models with better reasoning capabilities play a critical role in this task.

\begin{table}[h!]
\centering
\setlength{\tabcolsep}{12pt}
\begin{tabular}{c|lll}
VLM & R@1 & R@5 & R@10 \\ \hline
Without Re-ranking & 27.49 & 51.96 & 63.13 \\
Gemini 2.5 Flash Lite  & 23.54 & 48.39 & 63.13
  \\
Gemini 2.5 Flash  & 30.21 & 53.04 & 63.13
\end{tabular}%
\caption{VLM re-ranking comparison.
}
\label{tab:vlm_eval}
\vspace{-20pt}
\end{table}

\noindent\textbf{Ablation and Summary of \papername{}:} The ablation study of \papername{} and comparison with two previous state-of-the-art methods are stated in~\Cref{tab:sota}.
As summarised, \papername{} substantially outperforms two baselines~\cite{zheng2020university,wang2021each} on the University-1652 dataset. Also, the proposed drone-view augmentation and VLM re-ranking demonstrate their effectiveness in improving the performance on this benchmark.

\begin{table}[h!]
\centering
\setlength{\tabcolsep}{10pt}
\begin{tabular}{c|lll}
Model & R@1 & R@5 & R@10 \\ \hline
U1652~\cite{zheng2020university} & 1.20 & - & - \\
LPN w/o drone~\cite{wang2021each} & 0.74 & - & - \\
LPN w/ drone~\cite{wang2021each} & 0.81 & - & - \\ 
DINOv2-L & 24.66 & 48.00 & 59.02  \\
+ Drone Data & 27.49 & 51.96 & 63.13  \\
+ VLM Re-rank (Ours) & 30.21 & 53.04 & 63.13 
\end{tabular}%
\caption{Ablation study and baseline comparison.
}
\label{tab:sota}
\vspace{-20pt}
\end{table}


\section{CONCLUSION}
In this paper, we present \papername{}, a novel VLM-powered \ac{cvgl} model that achieved outstanding performance on the UAVM 2025 challenge~\cite{uavm2025} without introducing any extra datasets. To boost the coarse localisation performance of limited FOV query images, we introduce the drone perspective augmentation strategy. Furthermore, we prove that the reasoning capability of existing foundational \ac{vlms} significantly improves the localisation accuracy and provides fine-grained justifications for better interpretability, putting forward a new localisation paradigm for future research. 


\section*{ACKNOWLEDGEMENTS}
This work was supported by the National Science Foundation under Grants No. 2218063 and the EPSRC under grant agreement EP/S035761/1.
Computations were supported in part by Advanced Micro Devices, Inc., under the AMD AI \& HPC Cluster Program, and Vermont Advanced Computing Center (VACC).

{\small
\bibliography{egbib}

}


\end{document}